%% file: main_ieee.tex
\documentclass[letterpaper, 10 pt, conference]{ieeeconf}
\pdfoutput=1    % For LaTeX issues

\IEEEoverridecommandlockouts 

\usepackage{url}
\usepackage{cite}
\usepackage{amsmath,amssymb,amsfonts}
\usepackage{algorithmic}
\usepackage{graphicx}
\usepackage{textcomp}
\usepackage{xcolor}
\usepackage{booktabs}
\usepackage{subcaption}
\usepackage{pgfplots}
\usepackage{tikz}
\usetikzlibrary{spy}
\def\BibTeX{{\rm B\kern-.05em{\sc i\kern-.025em b}\kern-.08em
    T\kern-.1667em\lower.7ex\hbox{E}\kern-.125emX}}

\usetikzlibrary{svg.path}
\usepackage{scalerel}

\definecolor{orcidlogocol}{HTML}{A6CE39}
\tikzset{
  orcidlogo/.pic={
    \fill[orcidlogocol] svg{M256,128c0,70.7-57.3,128-128,128C57.3,256,0,198.7,0,128C0,57.3,57.3,0,128,0C198.7,0,256,57.3,256,128z};
    \fill[white] svg{M86.3,186.2H70.9V79.1h15.4v48.4V186.2z}
                 svg{M108.9,79.1h41.6c39.6,0,57,28.3,57,53.6c0,27.5-21.5,53.6-56.8,53.6h-41.8V79.1z M124.3,172.4h24.5c34.9,0,42.9-26.5,42.9-39.7c0-21.5-13.7-39.7-43.7-39.7h-23.7V172.4z}
                 svg{M88.7,56.8c0,5.5-4.5,10.1-10.1,10.1c-5.6,0-10.1-4.6-10.1-10.1c0-5.6,4.5-10.1,10.1-10.1C84.2,46.7,88.7,51.3,88.7,56.8z};
  }
}

\newcommand\orcidicon[1]{\href{https://orcid.org/#1}{\mbox{\scalerel*{
\begin{tikzpicture}[yscale=-1,transform shape]
\pic{orcidlogo};
\end{tikzpicture}
}{|}}}}

\usepackage{hyperref} %<--- Load after everything else
    
\begin{document}

\title{3D Point Cloud Compression with Recurrent Neural Network and Image Compression Methods*}

\author{Till Beemelmanns\,\textsuperscript{\orcidicon{0000-0002-2129-4082}$\dagger$}\,,
Yuchen Tao\,\textsuperscript{\orcidicon{0000-0002-0357-0637}$\dagger$}\,, 
Bastian Lampe\,\textsuperscript{\orcidicon{0000-0002-4414-6947}}\,, Lennart Reiher\,\textsuperscript{\orcidicon{0000-0002-7309-164X}}\,, \\
Raphael van Kempen\,\textsuperscript{\orcidicon{0000-0001-5017-7494}}\,,
Timo Woopen\,\textsuperscript{\orcidicon{0000-0002-7177-181X}}\,,  and Lutz Eckstein% <-this % stops a space
\thanks{*This research is accomplished within the project ”UNICARagil” (FKZ 16EMO0289). We acknowledge the financial support for the project by the Federal Ministry of Education and Research of Germany (BMBF).}% <-this % stops a space
\thanks{The authors are with the Institute for Automotive Engineering (ika), RWTH Aachen University, 52074 Aachen, Germany
        {\tt\small \{firstname.lastname\}@ika.rwth-aachen.de}}%
\thanks{$\dagger$ These authors contributed equally}%
}

%\author{\IEEEauthorblockN{Yuchen Tao\IEEEauthorrefmark{2}\textsuperscript{\textasteriskcentered},
%Till Beemelmanns\IEEEauthorrefmark{3}\textsuperscript{\textasteriskcentered},
%and Lutz Eckstein\IEEEauthorrefmark{3}}

%\IEEEauthorblockA{\IEEEauthorrefmark{2}RWTH Aachen, Germany\\
%yuchen.tao@rwth-aachen.de}
%\IEEEauthorblockA{\IEEEauthorrefmark{3}Insitute for Automotive Engineering, RWTH Aachen, Germany\\
%\{till.beemelmanns, lutz.eckstein\}@ika.rwth-aachen.de}}

%\author{Till Beemelmanns$^{1\dagger}$, Yuchen Tao$^{\dagger}$, and Lutz Eckstein$^{1}$% <-this % stops a space
%\thanks{*This research is accomplished within the project "UNICARagil" (FKZ16EMO0289). We acknowledge the financial support for the project by the Federal Ministry of Education and Research of Germany (BMBF).}% <-this % stops a space
%\thanks{$^{1}$The authors are with the Institute for Automotive Engineering (ika), RWTH Aachen University, 52074 Aachen, Germany
%        {\tt\small \{firstname.lastname\}@ika.rwth-aachen.de}}%
%\thanks{$\dagger$ These authors contributed equally}%
%}

%\maketitle
%\begingroup\renewcommand\thefootnote{\textasteriskcentered}
%\footnotetext{These authors contributed equally}
%\endgroup

\maketitle

\begin{abstract}
\input{sections/0_abstract}
\end{abstract}

%\begin{IEEEkeywords}
%Point Cloud Compression, Auto-Encoder, Recurrent Neural Network
%\end{IEEEkeywords}

\input{sections/1_introduction}
\input{sections/2_related_work}
\input{sections/3_method}
\input{sections/4_evaluation}
\input{sections/5_conclusion}
\input{sections/6_acknowledgment}

%%% REFERENCES
{\small
\bibliographystyle{IEEEtran}
\bibliography{main_ieee}
}

\end{document}

%% file: sections/0_abstract.tex
Storing and transmitting LiDAR point cloud data is essential for many AV applications, such as training data collection, remote control, cloud services or SLAM. However, due to the sparsity and unordered structure of the data, it is difficult to compress point cloud data to a low volume. Transforming the raw point cloud data into a dense 2D matrix structure is a promising way for applying compression algorithms. We propose a new lossless and calibrated 3D-to-2D transformation which allows compression algorithms to efficiently exploit spatial correlations within the 2D representation. To compress the structured representation, we use common image compression methods and also a self-supervised deep compression approach using a recurrent neural network. We also rearrange the LiDAR's intensity measurements to a dense 2D representation and propose a new metric to evaluate the compression performance of the intensity. Compared to approaches that are based on generic octree point cloud compression or based on raw point cloud data compression, our approach achieves the best quantitative and visual performance. Source code and dataset are available at \url{https://github.com/ika-rwth-aachen/Point-Cloud-Compression}.

%% file: sections/1_introduction.tex
\section{Introduction}
\label{sec:introduction}
With the development of automated vehicles (AVs), 3D LiDAR scanners have demonstrated their effectiveness for the perception of real world environments \cite{hecht2018lidar}. Laser scanners provide precise distant measurements even at bad illumination conditions and their application covers a variety of important tasks such as SLAM, object detection, point cloud segmentation and the detection of drivable area. As new laser scanner manufacturers enter the market, it can be expected that unit prices for LiDAR scanners will further decrease. Hence, we believe that LiDAR based sensors will continue to play an important role for future AVs.

Sharing and storing LiDAR point cloud data is necessary for many applications, such as training data collection, cooperative driving, cloud services or remote control \cite{UNICARagil}. However, LiDAR sensors sample several thousand data points for each data frame, which results in a large data size for a single LiDAR scan. For example, a single VLP-32C sensor produces around 1 Gigabyte of data every minute. This data size causes a high transmission latency and leads to a big data storage footprint.

Point cloud data is difficult to process and compress due to its sparsity and unordered structure. A point cloud is usually represented by a list of coordinates and additional point features. Data compression which aims at reducing redundancy in the data is therefore difficult to apply. Hence, it is a popular approach to first transform the point cloud into a structured representation, such as tree structures, height maps, voxels or 2D representations. Then, generic compression approaches can be applied which exploit the spatial correlations within these representations \cite{octree, rusu20113d, draco, heightmap, wavelet, gollarealtime, tu2016compressing, tuRNN, UnetTu, panorama, large, sun2019novel, sun2020novel}. Recent approaches also make use of deep neural networks, designed for image compression, for the compression of raw point cloud data \cite{tuRNN}. \\

In this paper, we focus on point cloud compression and propose a compression approach that projects a LiDAR scan to a combination of calibrated range, azimuth and intensity representations. Then, we use image compression methods and a deep recurrent neural network model to compress these representations.

The main contributions of this paper are as follows:
\begin{itemize}
    \item Novel calibrated and lossless projection of a point cloud into several image-like representations.
    \item Compression of these representations with image compression and a self-supervised deep compression approach.
    \item Comparison with reference methods using a novel evaluation metric which pays special attention to the compressibility of the intensity.
\end{itemize}
Our results show that our calibrated point cloud transformation allows common generic image compression methods and a deep compression approach to achieve better results than comparable methods. We share our source code and datasets to reproduce the results presented in this work.

%% file: sections/2_related_work.tex
\section{Related Work}
\label{sec:related_work}
Point cloud data is particularly challenging to process and compress due to its sparsity and disordered structure. Therefore, it is common to transform the data into a structured representation and then apply compression algorithms.

A point cloud can be transformed into a \emph{tree-based} structure by partitioning the points into separate volumes. The two most common tree representations of point clouds are the k-d tree and the octree. The tree structures can then be used for lossy compression \cite{octree, rusu20113d, draco}. The compression is achieved by replacing points in a tree's leaves with the cell centers and quantizing the coordinates of the points \cite{octree}. Tree structures are also used to develop more complex point cloud compression algorithms. For example, Schnabel and Klein \cite{octree} designed a lossless compression method especially suited for densely sampled point clouds, based on an octree space decomposition. All quantized leaf center coordinates are efficiently encoded by predicting which cells are occupied. The popular Point Cloud Library (PCL) supports octree compression \cite{rusu20113d}, and k-d tree based point cloud compression is implemented in the Google Draco project \cite{draco}.

Pauly and Gross \cite{heightmap} first came up with the idea to represent point clouds as \emph{height maps} by splitting the point cloud into patches, which are then mapped to a scalar height field tessellation representation. Finally, the patches are resampled using a regular grid and processed by a Discrete Fourier Transform (DFT) to obtain a spectral representation. The spectral representation can be manipulated to achieve different effects. Inspired by this work, Ochotta and Saupe \cite{wavelet} first segment a point cloud into patches, map the patches to base planes to get height field maps, then used a shape-adaptive wavelet coder to encode the height maps in order to compress the point cloud. Golla and Klein \cite{gollarealtime} introduced a simple decomposition of the point cloud into patches by simply segmenting the point cloud into voxels and taking each voxel as a patch. Then, standard image compression methods were applied to these height maps.

Another popular approach to compress point cloud data is to transform the point cloud to a \emph{2D image representation} and then apply image compression algorithms \cite{tu2016compressing, tuRNN, UnetTu, panorama, large, sun2019novel, sun2020novel}. Tu et al. \cite{tu2016compressing} used raw packet data from Velodyne laser scanners and converted them into 2D images. Then, conventional image compression methods were applied to the image sequences. In \cite{tuRNN, UnetTu}, machine learning-based image and video compression methods were applied to compress the image sequences. Other approaches focus on the compression of point clouds obtained by terrestrial laser scanners \cite{panorama, large}.  Houshinar and Nüchter \cite{panorama} projected the range, color, and reflectance information of a point cloud onto camera images, then investigated the efficiency of conventional image compression methods on the fused representation. Hubo et al. \cite{large} used a similar projection approach and proposed a hybrid range image coding algorithm that predicts a pixel using previously encoded neighbor pixel values. Sun et al. \cite{sun2019novel} performed segmentation on a range image converted from a point cloud. To reduce redundancy in the image, a predictive compression method was applied on the segmented clusters. In their following work, LSTM cells are introduced to predict the inter-frames differences to remove spatial redundancy \cite{sun2020novel}.

A different approach to compress point clouds would be to transform the points into other structured representations like \emph{voxels}, then encode them with an auto-encoder \cite{quachPCGC, syndrome}. Inspired by JPEG compression which encodes 2D images in $8\times8$ blocks, Chen et al. proposed a transformation named Point Cloud Neural Transform (PCT) \cite{pct} to encode 3D large-scale point clouds into voxels. PCT achieved good compression results in the setting of automated driving. The irregularity of points in each voxel is eliminated by building a spatial graph, where each point is connected to its K nearest neighbors.

%% file: sections/3_method.tex
\section{Method}
\label{sec:method}
\begin{figure}[h!]
    \includegraphics[width=1\linewidth]{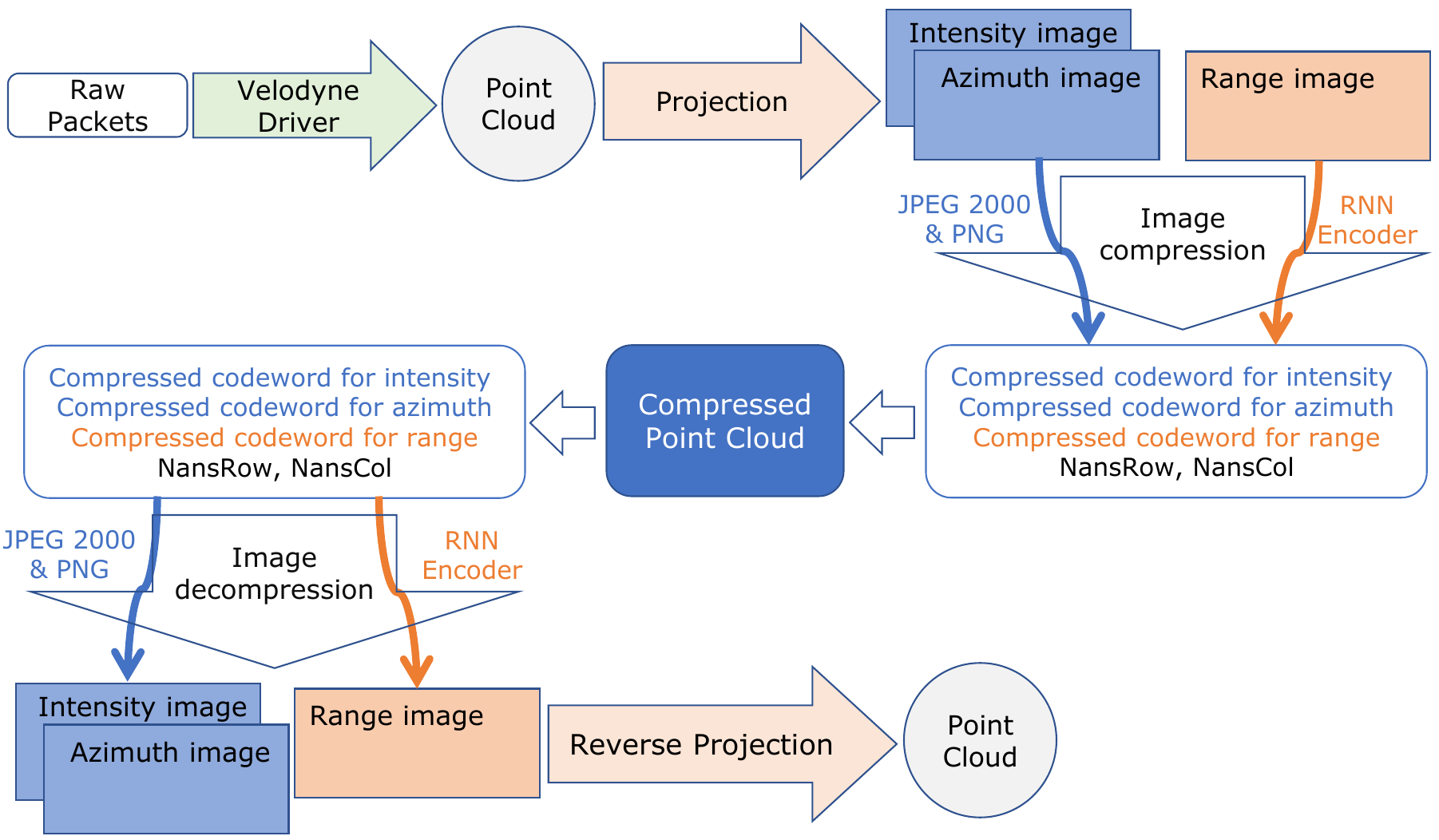}
    \centering
    \caption{Flowchart of the proposed compression method.}
    \label{fig:detailpipeline}
\end{figure}

The overall approach of this paper can be divided into three steps, as illustrated in Fig. \ref{fig:detailpipeline}. In the first step, the point cloud is projected losslessly into three 2D matrix representations. The range image, the azimuth image and the intensity image. We preprocess these representations with a novel approach and send them to the encoder-compressor. The encoder performs image compression using a recurrent neural network and traditional image compression methods, and it generates a codeword for each representation. The codewords are combined into a single message, which is then transmitted to the decoder. The decoder decompresses the codewords and assembles the reconstructed point cloud.

In contrast to related work \cite{ImageRNN}, we propose an improved way to construct the 2D matrix representations which allows the RNN-based and the image compression methods to exploit spatial correlations much better. Furthermore, we explicitly compress the intensity field of the point cloud, which has not been covered by literature yet. We also propose a metric to evaluate the compression of the intensity and we suggest a new method to evaluate the geometric compression quality.    

\subsection{Point Cloud Data Transformation}
\subsubsection{Range Image Projection}
\begin{figure}[h!]
  \centering
    \includegraphics[width=0.9\linewidth]{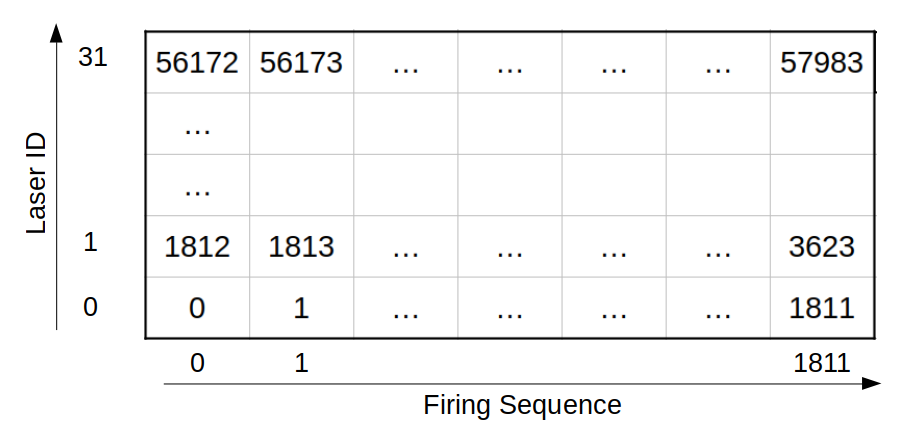}
  \caption{The 2D structure of a LiDAR scan. The sequence numbers from 0 to 57983 indicate in which order the points are stored within a single message.}
  \label{fig:pointcloud2}
\end{figure}

%For each point in the 3D space, the distance from the LiDAR sensor to the point can be %calculated from its $(x,y,z)$ coordinates by
%\begin{equation}
%d=\sqrt{(x^2+y^2+z^2)}.
%\end{equation}
%
%To simplify the implementation, the $32\times1800$ range images are also stored with shape $32\times1812$, where the last $12$ columns are regarded as \texttt{NaN} value points.
In one $360^\circ$ revolution of our Velodyne VLP-32C laser scanner, each laser channel samples the same number of data points including the \texttt{NaN} values, where no reflection is received. That means we obtain a 2D representation of the point cloud as shown in  Fig. \ref{fig:pointcloud2}. We can convert a single sensor scan into a $32\times1812$ range image, where each pixel contains the distance from the sensor to the corresponding point. Note that we use in our setting a sensor frequency of 10Hz. Each row of the range image corresponds to one laser channel of our laser scanner and each column corresponds to the firing sequence. The float distance value $d$ calculated from the $(x,y,z)$ coordinates is converted to a 16-bit integer value by
\begin{equation}
 d_{\text{pixel}}=\lfloor\frac{d}{d_{max}}\times65535\rfloor 
\end{equation}
with $d_{max}=200\;m$ as the maximum range of the VLP-32C model, and $65535$ as the maximum value of a 16-bit integer.

\subsubsection{Azimuth Offset Shifting}
\begin{figure}[h!]
  \centering
  \begin{subfigure}[b]{1\linewidth}
      \includegraphics[width=\linewidth]{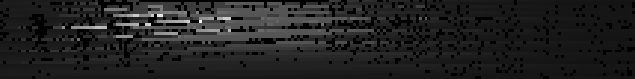}
  \caption{Visualization of the naive range image. Zoomed in.}\label{fig:naive1}
  \end{subfigure}
  \begin{subfigure}[b]{1\linewidth}
      \includegraphics[width=\linewidth]{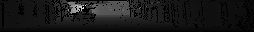}
  \caption{The corresponding snippet after shifting. Zoomed in.}\label{fig:naive2}
  \end{subfigure}
  \caption{Shifting the rows of the naive range image according to the azimuth offsets.}
  \label{fig:naive}
\end{figure}

As each laser channel has an azimuth offset, the points in each column of the 2D matrix do not share the same azimuth angle, as shown in Fig. \ref{fig:naive1}. Thus, each row of the resulting range image can be shifted according to this offset to align pixels in each column as shown in Fig. \ref{fig:naive2}. The specific offset for each laser channel depends on the position of the laser array and is determined by calibration. 

\subsubsection{Range Image Denoising}
\begin{figure}[h!]
  \centering
  \begin{subfigure}[b]{0.95\linewidth}
      \includegraphics[width=\linewidth]{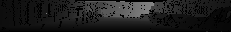}
  \caption{The range image with dropout noise.}\label{fig:denoising1}
  \end{subfigure}
  \begin{subfigure}[b]{0.95\linewidth}
      \includegraphics[width=\linewidth]{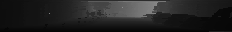}
  \caption{The range image after the denoising processing step.}\label{fig:denoising2}
  \end{subfigure}
  \caption{The effect of denoising for the range image.}
  \label{fig:denoising}
\end{figure}

We can see from Fig. \ref{fig:naive2} that there is a lot of dropout noise in the range image visualization. This noise, colored as black pixels, represents \texttt{NaN} values of the point cloud. \texttt{NaN} is returned from the laser scanner if no reflectance is received. This dropout noise substantially affects the compression quality because it is a random high-frequency component that is hard to compress. Hence, we propose to fill these \texttt{NaN} values so that the representations becomes smoother. This is realized as follows: during the conversion, once a \texttt{NaN} point is encountered, its position indexes are stored into two arrays, an integer array storing the row index and another integer array which stores the column index. This additional information only takes up a few kilobytes and is sent along with the compressed point cloud data for the reverse projection. The \texttt{NaN} values in the image are filled with the previous pixel values. The result of this denoising step is shown in Fig. \ref{fig:denoising}.

\subsubsection{Azimuth Image Assembly}
\begin{figure}[h!]
  \centering
    \begin{subfigure}[b]{1\linewidth}
    \includegraphics[width=1\linewidth]{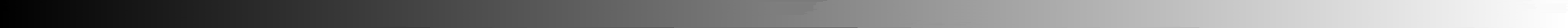}
    \caption{Azimuth image starting from $-180^\circ$(left) to $180^\circ$(right).}
    \end{subfigure}
    \begin{subfigure}[b]{1\linewidth}
    \includegraphics[width=1\linewidth]{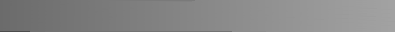}
    \caption{Azimuth image. Zoomed in.}
    \end{subfigure}
  \caption{Example of an azimuth image.}
  \label{fig:azimuth}
\end{figure}

The generation of the azimuth images can be done simultaneously while the range images are generated. To calculate the azimuth, the point coordinates in the world coordinate $(x,y,z)$ are projected to the range image system, then the azimuth is calculated by $arctan(\frac{y}{x})$ and stored as a 16-bit integer. The azimuth image representation is shown in Fig. \ref{fig:azimuth}. It is a very smooth image with little noise, making it easy to compress with common image compression methods because it contains only few high-frequency components.

\subsubsection{Intensity Image Assembly}
\begin{figure}[h!]
  \centering
  \begin{subfigure}[b]{1\linewidth}
    \includegraphics[width=\linewidth]{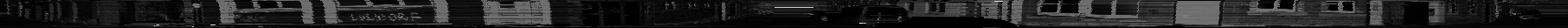}
    \caption{The intensity image.}
  \end{subfigure}
  \begin{subfigure}[b]{1\linewidth}
    \includegraphics[width=\linewidth]{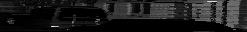}
    \caption{Zoomed intensity image.}
  \end{subfigure}
  \caption{Example for an intensity 2D image representation.}
  \label{fig:intensity}
\end{figure}

The intensity represents the reflectivity of the object's surface and is an important feature for numerous applications such as point cloud classification and object detection. However, few pieces of research focus on compressing the intensity values along with the coordinates. Within this work, an intensity image is generated from each point cloud in the same way as the range image and azimuth images. Because the intensities are only 8-bit integers, the raw intensity images are half the size of the raw range images and azimuth images. A visualization of the intensity representation is shown in Fig. \ref{fig:intensity}.
 
\subsubsection{Point Cloud Reconstruction}\label{sec:reconstruction}
After the transformation to the range, azimuth, and intensity representations, these three 2D matrices along with the information of the \texttt{NaNs} positions can be used to reconstruct the original point cloud losslessly.

The $x$, $y$, and $z$ coordinates can be retrieved using the distance $d$, the azimuth angle $\alpha$, and the elevation angle $\omega$. The elevation angle can be determined from the row ID of the data point in the range image, because each laser channel has a specific elevation angle, which can be looked up from the sensor model configuration. Then, the final coordinates can be calculated by
\begin{align}
    \begin{split}
    x&=d*sin(-\alpha)*cos(\omega), \\
    y&=d*cos(\alpha)*cos(\omega), \\
    z&=d*sin(\omega).
    \end{split}
\end{align}
Points that correspond to the stored \texttt{NaN} values are skipped.% The reconstructed point cloud compared to the original point cloud has a SNNRMSE of around $0.002$ m, due to the loss of accuracy using floating-point operations. This is a negligible error so that we can regard this projection method as lossless.

\subsection{Evaluation of Compressibility}
To evaluate the compressibilities of the three representations, we randomly sampled a subset of LiDAR scans from our dataset and transformed them into range, intensity, and azimuth images. Then, PNG compression was applied to the images, and an average compression rate for each type of image was calculated. The result of the experiment is shown in Table \ref{tab:png}.
\begin{table}[h!]
    \begin{center}
        \begin{tabular}{cccc}
            \toprule 
            \textbf{Representation} & \textbf{Range} & \textbf{Intensity} & \textbf{Azimuth}\\
            \midrule 
            Compression Rate & 55.34$\%$ & 49.32$\%$ & 29.36$\%$\\
            \bottomrule
        \end{tabular}
    \end{center}
    \caption{Compression rates of the range, intensity, and azimuth images with PNG compression.}\label{tab:png}
\end{table}%
As indicated in the table, the range and intensity images are difficult to compress, while the azimuth images are relatively easy to compress. This can be explained by the fact that the spatial correlation between pixels in the range image and intensity image is weaker compared to the azimuth image. There are more sharp edges and high frequency information compared to the azimuth image. This compressibility test helped us choose the most efficient compression method for the three representations for our overall compression approach.

\subsection{Compression Approach}
\subsubsection{Compression of Azimuth}
The original azimuth image representation has a size of $116$ Kilobytes. Compressing it with JPEG 2000 to around $11$ Kilobytes introduced a negligible small error on our test dataset. The compression of the azimuth representation with PNG compression did not introduce any error in the reconstructed point cloud, but it could only achieve a compression rate of $29.36\%$. Hence, we favored the usage of JPEG 2000 which achieves a compression rate of $10\%$ and induces only a negligible error in the context of lossy point cloud compression.

\subsubsection{Compression of Intensity}
According to the compressibility test in Table \ref{tab:png}, the intensity representation is harder to compress than the azimuth images. However, the raw intensity image is only half the size of the range and azimuth images, as it is an 8-bit image. Hence, we used traditional image compression algorithms to compress the intensity representation, as it is not the bottleneck of the compression task. We evaluated PNG, JPEG and JPEG 2000 compression on a set of intensity images, and we used our novel $\text{SNNRMSE}_\text{I}$ metric (cf. Eq. \ref{eq:snnrmsei}) to compare the compression methods. We observed that JPEG 2000 performs much better than JPEG. However, we chose to use lossless PNG compression, which compresses the intensity representation from $58$ Kilobytes to around $28$ Kilobytes.

\subsubsection{Compression of Range}
\begin{figure}[h!]
  \centering
    \includegraphics[width=1\linewidth]{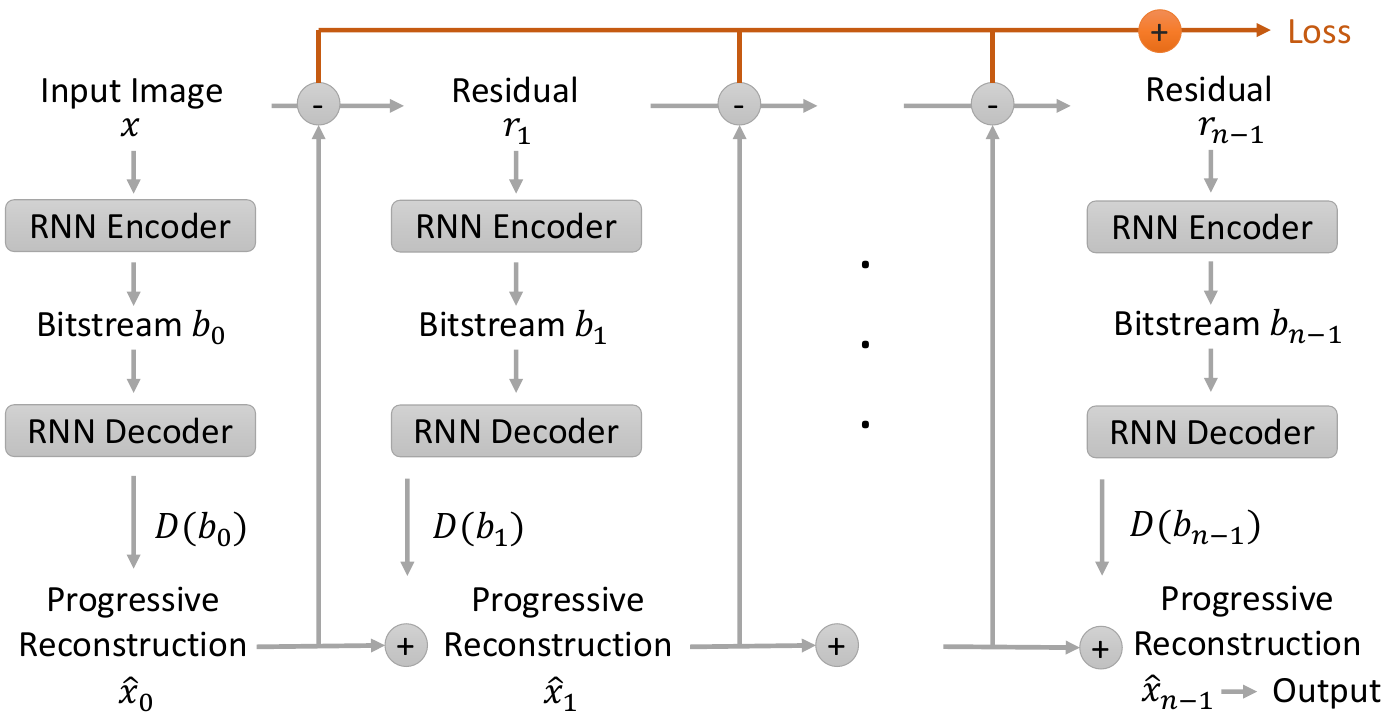}
  \caption{The additive RNN reconstruction framework.}
  \label{fig:additive}
\end{figure}

As the range representation is difficult to compress, we used a deep compression approach based on a RNN architecture which has proven to work well on natural image compression. This compression model is based on the work by Toderici et al. \cite{ImageRNN} and Tu et al. \cite{tuRNN}. The RNN consists of an Encoder-Decoder pair with a Binarizer in between both network parts. A single iteration of the RNN can be represented by
\begin{align}
 \begin{split}
b_t &= B(E_t(r_{t-1})), \\
\hat{x}_t &= D_t(b_t)+\hat{x}_{t-1}, \\
r_0&=x,\\
r_t&=x-\hat{x}_t, \\
\hat{x}_0&=0, \\
 \end{split}
\end{align}
where $t$ denotes the current iteration of the RNN, $b_t$ is a $m$ bits binarized stream, $b_t\in\{-1,1\}^m$, $B$ is the Binarizer, $E_t$ and $D_t$ represents the Encoder and Decoder with their cell states at iteration $t$, $r_t$ is the difference between the progressive reconstruction $\hat{x}_t$ and the original input $x$. With a total number of $n$ iterations, the loss of the RNN model is calculated by 
\begin{equation}
 L=\frac{1}{n}\sum_t\lvert{r_t}\rvert. 
\end{equation}
In each iteration the network reconstructs the residuals from the last iteration, and the final output $\hat{x}_{n-1}$ is progressively reconstructed by adding the newly predicted residuals, as illustrated in Fig. \ref{fig:additive}. 

We also evaluated a \emph{one-shot reconstruction} architecture proposed by Toderici et al. \cite{ImageRNN}, which achieved promising performance for natural image compression. A comparison  revealed that it performed equally or slightly worse than the additive framework in our setting.
Furthermore, we evaluated the impact on the compression performance for additional GND layers proposed by Islam et al. \cite{Islam_2021_CVPR}. We incorporated two GND layers after the first convolutional layer of encoder and decoder and also evaluated different data normalization methods, but we did not observe any improvements in the performance.

\subsubsection{Range Image Normalization}
Although the Velodyne VLP-32C model provides a maximum range of $200$\ m, most of the points it detects are within a much shorter range. We randomly sampled a subset of frames from our dataset and computed the mean of the range for each point. We normalized the range image's pixel values to $[0,1]$ and further normalized the scaled image $R$ with $(R-\mu)/\theta$, where $\mu$ is the average range, and $\theta$ is the $95^{th}$-percentile of all ranges. The normalized range images are used to train the additive RNN model. 

\subsubsection{Training Protocol}
Instead of directly training on the $32\times1812$ range images, the image is randomly cropped to $32\times32$ batches. During prediction, the image is padded to make sure it will be divisible by $32$.
We use a dataset of ten distinct sequences that contain raw Velodyne scan packets. The sequences are acquired across different areas in Aachen. A single sequence containing $1217$ frames is used as the test dataset and the evaluation in this paper are performed on a fixed subset of frames from this sequence. The training dataset contains $30813$ frames. The remaining frames are used as validation dataset. We train the model for 2000 epochs using a cosine learning scheduler.

% During training, the cosine learning scheduler \cite{loshchilov2016sgdr} was applied, and it has proven to be a useful strategy.

%% file: sections/4_evaluation.tex
\section{Evaluation}
\label{sec:evaluation}
\subsection{Reference Methods}
To compare with the proposed model, the following four methods are used as reference. 
\begin{enumerate}
    \item \emph{PCL Octree} \cite{rusu20113d}: PCL provides point cloud compression based on the octree structure. The framework does not support the compression of points that carry the additional intensity field. Hence, we modified the code and stored the intensity value into the R channel of the point type \emph{XYZRGB}.
    \item \emph{Google Draco} \cite{draco}: Google Draco is a library for compressing 3D geometric meshes and point clouds. It supports k-d tree based point cloud compression and the intensity field compression. 
    \item \emph{Tu et al.} \cite{tu2016compressing}: In this approach, the raw Velodyne scan data is first projected to a range image and to an intensity image, then compressed using image compression methods.
    \item \emph{Proposed Method using JPEG 2000}: Since JPEG 2000 has shown its power in compressing azimuth and intensity images, we can also use it to compress the range image representation.
\end{enumerate}

\subsection{Evaluation metrics}
The metric Symmetric Nearest Neighbor Root Mean Squared Error (SNNRMSE) \cite{gollarealtime} is used to evaluate the difference between two point clouds. Given two point clouds $P$ and $Q$, each point $p\in{P}$ can find the nearest neighbor $q\in{Q}$. The RMSE is defined by: 
\begin{equation}
\text{RMSE}_{\text{NN}}(P,Q) = \sqrt{\sum_{p\in{P}}(p-q)^2/\lvert P \rvert}, 
\end{equation}
where $\lvert P \rvert$ is the number of points in $P$. To derive a symmetric metric, the $\text{SNNRMSE}$ is defined as: 
\begin{equation}
\begin{aligned}
& \text{SNNRMSE}(P,Q) =  \\
& \sqrt{0.5*\text{RMSE}_\text{NN}(P,Q)+0.5*\text{RMSE}_\text{NN}(Q,P)}. 
\end{aligned}
\end{equation}
We also propose the new metric $\text{SNNRMSE}_\text{I}$ to measure the difference of intensity values between two point clouds: 
\begin{equation}
\begin{aligned}
& \text{SNNRMSE}_\text{I}(P,Q) = \\ 
& \sqrt{0.5*\text{RMSE}_\text{I}(P,Q)+0.5*\text{RMSE}_\text{I}(Q,P)}, 
\end{aligned}
\label{eq:snnrmsei}
\end{equation}
where $\text{RMSE}_\text{I}$ is defined by 
\begin{equation}
\text{RMSE}_{\text{I}}(P,Q) = \sqrt{\sum_{p\in{P}}(p.I-q.I)^2/\lvert P \rvert}, 
\end{equation}
where $p.I$ and $q.I$ are the intensity values of point $p$ and $q$.

Furthermore, we measured the size of a compressed point cloud by computing the metric bpp (bits per point), which indicates how many bits are used to encode one point in the point cloud on average. In particular, we explicitly measure the whole compressed point cloud including the intensity values and divide it through the number of points in the point cloud.

\subsection{Quantitative evaluation}

\subsubsection{Distortion}
Similar to related work \cite{tuRNN}, we computed the average SNNRMSE between the original point clouds and the decompressed point clouds against the average $\text{bpp}$. To keep the comparison fair, all point cloud compression methods compress the intensity values of the point cloud losslessly. Note that an original point cloud has a memory footprint of 196 $\text{bpp}$. Transforming this point cloud into the uncompressed image representations along with the stored \texttt{NaN} values reduces this footprint to 48 $\text{bpp}$.

\begin{figure}[h!]
\centering
\begin{tikzpicture}
\pgfplotstableread{data/intensity_distortion_draco.dat}{\intensitydisdraco}
\pgfplotstableread{data/intensity_distortion_octree.dat}{\intensitydisoctree}
\pgfplotstableread{data/intensity_distortion_proposed.dat}{\intensitydisproposed}
\pgfplotstableread{data/intensity_distortion_proposed_rnn.dat}{\intensitydisproposedrnn}
\pgfplotstableread{data/intensity_distortion_tu.dat}{\intensitydistu}
\begin{axis}[
    xmin = 9, xmax = 35,
    ymin = 0, ymax = 0.25,
    xtick distance = 5,
    ytick distance = 0.05,
    grid = both,
    minor tick num = 1,
    major grid style = {lightgray},
    minor grid style = {lightgray!25},
    width = \linewidth,
    height = 0.75\linewidth,
    legend cell align = {left},
    legend pos = north east,
    legend style={font=\scriptsize},
    xlabel={Bits per point [bpp]},
    ylabel={SNNRMSE [m]},
    style={/pgf/number format/fixed}
]
\addplot[blue, mark = *] table [x = {bpp}, y = {SNNRMSE}] {\intensitydisdraco};
\addplot[red, mark=square*] table [x ={bpp}, y = {SNNRMSE}] {\intensitydisoctree};
\addplot[teal, mark = x, mark size = 3pt] table [x = {bpp}, y = {SNNRMSE}] {\intensitydistu};
\addplot[green, mark = triangle*, mark size = 3pt] table [x = {bpp}, y = {SNNRMSE}] {\intensitydisproposed};
\addplot[cyan, mark = diamond*, mark size = 3pt] table [x = {bpp}, y = {SNNRMSE}] {\intensitydisproposedrnn};
\legend{
    Google Draco, 
    PCL Octree,
    Tu et al. using JPEG 2000,
    Proposed Method using JPEG 2000,
    Proposed Method using RNN
}
\end{axis}
\end{tikzpicture}
\caption{Evaluation of compression performance.}
\label{fig:quality}
\end{figure}
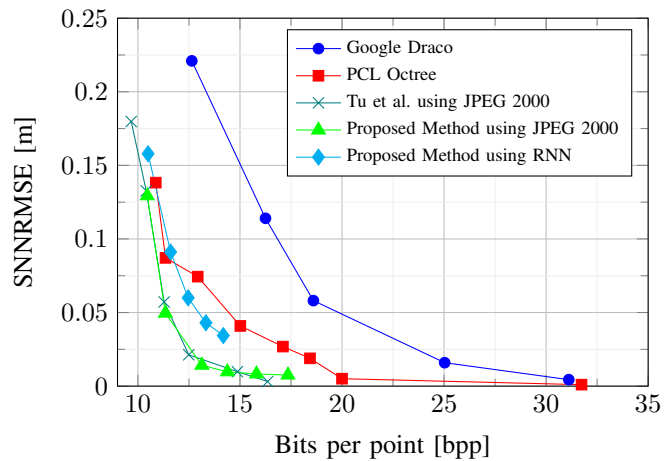

% For k-d tree and octree, we encoded the intensity with full 8-bit resolution. And for the model proposed by Tu et al. and our proposed model, it means compressing the intensity images with lossless PNG compression. To avoid point loss, the grid downsampling option for tree-based compression methods is disabled. For Google Draco, we used the lowest coding speed to ensure the best compression performance of k-d tree compression. 
With respect to the SNNRMSE, our proposed methods perform similar or worse than the reference method by Tu et al., as shown in Fig. \ref{fig:quality}.
However, we think that this metric is not fully suitable to measure the performance of the compression method. The SNNRMSE and the observable distortion of the reconstructed point cloud do not correlate well (cf. Fig \ref{fig:distortion}). This behavior can also be observed in the work by Tu et al. \cite{tuRNN}. The SNNRMSE seems to be very sensitive to single outliers and therefore the overall performance evaluation of the whole point cloud is not very meaningful.  

\begin{figure}[h!]
\centering
\begin{tikzpicture}
\pgfplotstableread{data/intensity_distortion_draco.dat}{\intensitydisdraco}
\pgfplotstableread{data/intensity_distortion_octree.dat}{\intensitydisoctree}
\pgfplotstableread{data/intensity_distortion_proposed.dat}{\intensitydisproposed}
\pgfplotstableread{data/intensity_distortion_proposed_rnn.dat}{\intensitydisproposedrnn}
\pgfplotstableread{data/intensity_distortion_tu.dat}{\intensitydistu}
\begin{axis}[
    xmin = 9, xmax = 35,
    ymin = -0.5, ymax = 25,
    xtick distance = 5,
    ytick distance = 5,
    grid = both,
    minor tick num = 1,
    major grid style = {lightgray},
    minor grid style = {lightgray!25},
    width = \linewidth,
    height = 0.75\linewidth,
    legend cell align = {left},
    legend pos = north east,
    legend style={font=\scriptsize},
    xlabel={Bits per point [bpp]},
    ylabel={Intensity SNNRMSE [I]},
]
\addplot[blue, mark = *] table [x = {bpp}, y = {SNNRMSEI}] {\intensitydisdraco};
\addplot[red, mark=square*] table [x ={bpp}, y = {SNNRMSEI}] {\intensitydisoctree};
\addplot[teal, mark = x, mark size = 3pt] table [x = {bpp}, y = {SNNRMSEI}] {\intensitydistu};
\addplot[green, mark = triangle*, mark size = 3pt] table [x = {bpp}, y = {SNNRMSEI}] {\intensitydisproposed};
\addplot[cyan, mark = diamond*, mark size = 3pt] table [x = {bpp}, y = {SNNRMSEI}] {\intensitydisproposedrnn};
\legend{
    Google Draco, 
    PCL Octree,
    Tu et al. with JPEG 2000,
    Proposed Method using JPEG 2000,
    Proposed Method using RNN
}
\end{axis}
\end{tikzpicture}
\caption{Distortion estimation by evaluating $\text{SNNRMSE}_\text{I}$ induced by spatial distortion of the point cloud.}
\label{fig:intensitydistortion}
\end{figure}
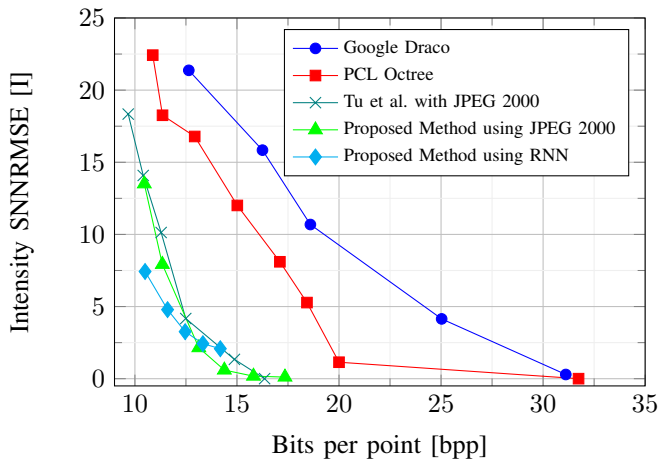

Hence, we evaluated the distortion also from another perspective by comparing the $\text{SNNRMSE}_\text{I}$ of the different methods. Since the intensity values are losslessly compressed, the increase in $\text{SNNRMSE}_\text{I}$ is only introduced by the change of nearest neighbors due to distortion. Because near points in a point cloud usually have similar intensity values, a higher $\text{SNNRMSE}_\text{I}$ indicates that points in the reconstructed point cloud may be shifted further away. Fig. \ref{fig:intensitydistortion} shows that our RNN image compression model resulted in the minimum $\text{SNNRMSE}_\text{I}$ at a very low compression rate, and the increase in $\text{SNNRMSE}_\text{I}$ for higher compression rates is slower compared to Tu et al. and the proposed method with JPEG 2000.

\subsubsection{Intensity}
To evaluate the compression of intensity values, the geometric information is not compressed. We can evaluate the compression rate for intensity by calculating the average bit used to represent the intensity value for one point in a point cloud. For Google Draco and PCL octree compression, we can directly set how many bits we want to encode intensity values, and for the method by Tu et al. and our proposed method, this can be calculated by the size of the compressed range image divided by the number of points.
\begin{figure}[h!]
\centering
\begin{tikzpicture}
\pgfplotstableread{data/intensity_plot_draco.dat}{\intensitydatadraco}
\pgfplotstableread{data/intensity_plot_octree.dat}{\intensitydataoctree}
\pgfplotstableread{data/intensity_plot_proposed.dat}{\intensitydataproposed}
\pgfplotstableread{data/intensity_plot_tu.dat}{\intensitydatatu}
\begin{axis}[
    xmin = 0, xmax = 8,
    ymin = 0, ymax = 30,
    xtick distance = 1,
    ytick distance = 5,
    grid = both,
    minor tick num = 1,
    major grid style = {lightgray},
    minor grid style = {lightgray!25},
    width = \linewidth,
    height = 0.75\linewidth,
    legend cell align = {left},
    legend pos = north east,
    legend style={font=\scriptsize},
    xlabel={Average bits per intensity value [bit]},
    ylabel={Intensity SNNRMSE [I]},
]
\addplot[blue, mark = *] table [x = {bit}, y = {draco}] {\intensitydatadraco};
\addplot[red, mark=square*] table [x ={bit}, y = {octree}] {\intensitydataoctree};
\addplot[teal, mark = x, mark size = 3pt] table [x = {bit}, y = {tu}] {\intensitydatatu};
\addplot[green, mark = triangle*, mark size = 3pt] table [x = {bit}, y = {proposed}] {\intensitydataproposed};
\legend{
    Google Draco, 
    PCL Octree,
    Tu et al. using JPEG 2000,
    Proposed Method using JPEG 2000
}
\end{axis}
\end{tikzpicture}
\caption{Intensity compression quality using four different methods.}
\label{fig:intensityevaluation}
\end{figure}
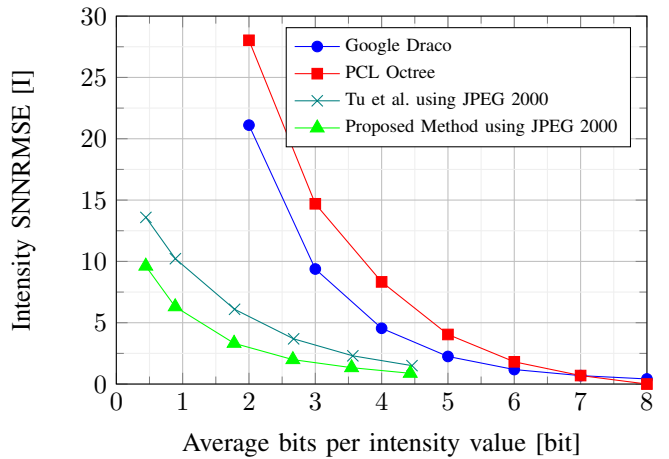
Fig. \ref{fig:intensityevaluation} shows that using JPEG 2000 to compress the intensity image in our proposed method achieves the best compression results for intensity. The result of compressing the intensity images by Tu et al. is worse than ours because their intensity images are raw and uncalibrated, making them harder to compress.

\subsection{Qualitative Evaluation}
\begin{figure*}[h]
  \centering
    \begin{subfigure}[t]{0.32\linewidth}
    \resizebox{\linewidth}{!}{%
    \begin{tikzpicture}[spy using outlines={circle,yellow,magnification=2.5,size=5cm, connect spies}]
    \node {\pgfimage{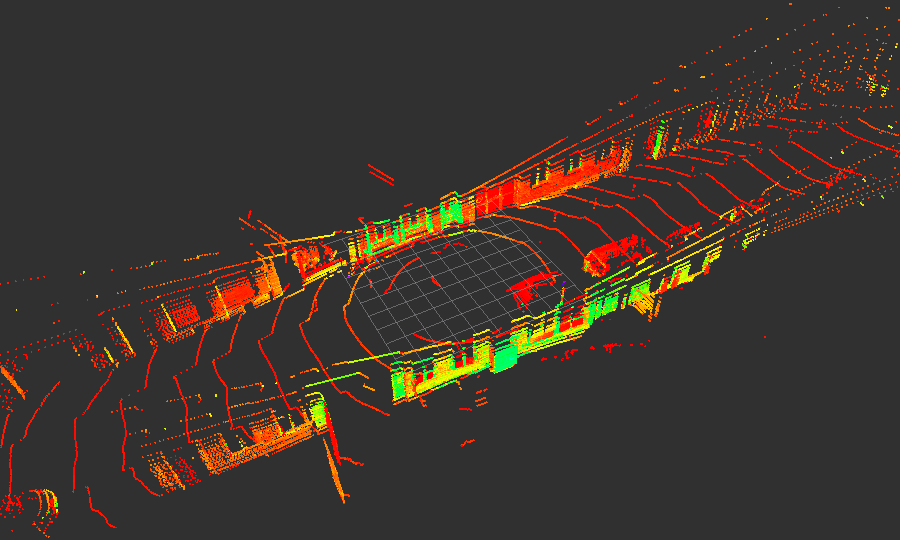}};
    \spy on (4.3,0.4) in node [right] at (-11,4.0);
    \end{tikzpicture}%
    }
    \caption{Original Point Cloud, \\ SNNRMSE=0.0 m.}\label{fig:distortion0}
  \end{subfigure}
  \begin{subfigure}[t]{0.32\linewidth}
    \resizebox{\linewidth}{!}{%
    \begin{tikzpicture}[spy using outlines={circle,yellow,magnification=2.5,size=5cm, connect spies}]
    \node {\pgfimage{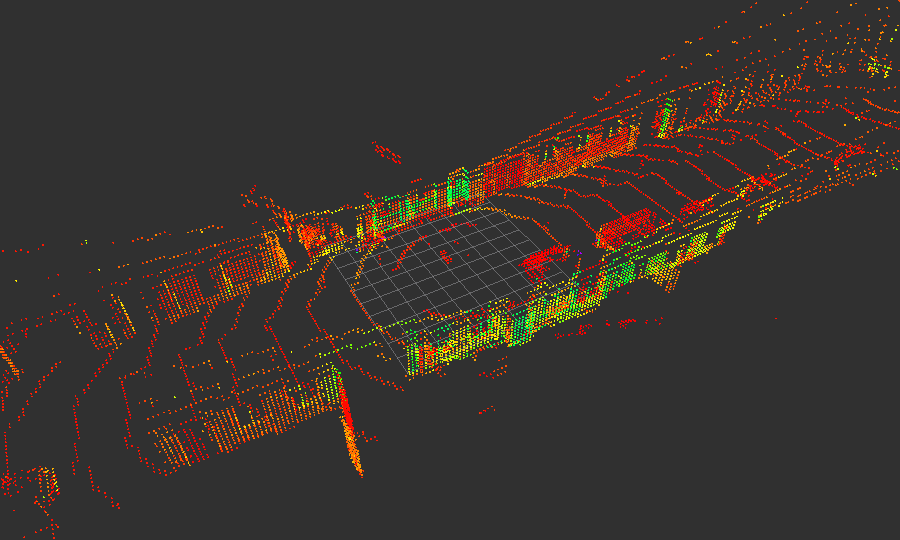}};
    \spy on (4.5,1.0) in node [right] at (-11,4.0);
    \end{tikzpicture}%
    }
    \caption{PCL Octree Compression, \\ SNNRMSE=0.1206 m.}\label{fig:distortion1}
  \end{subfigure}
  \begin{subfigure}[t]{0.32\linewidth}
    \resizebox{\linewidth}{!}{%
    \begin{tikzpicture}[spy using outlines={circle,yellow,magnification=2.5,size=5cm, connect spies}]
    \node {\pgfimage{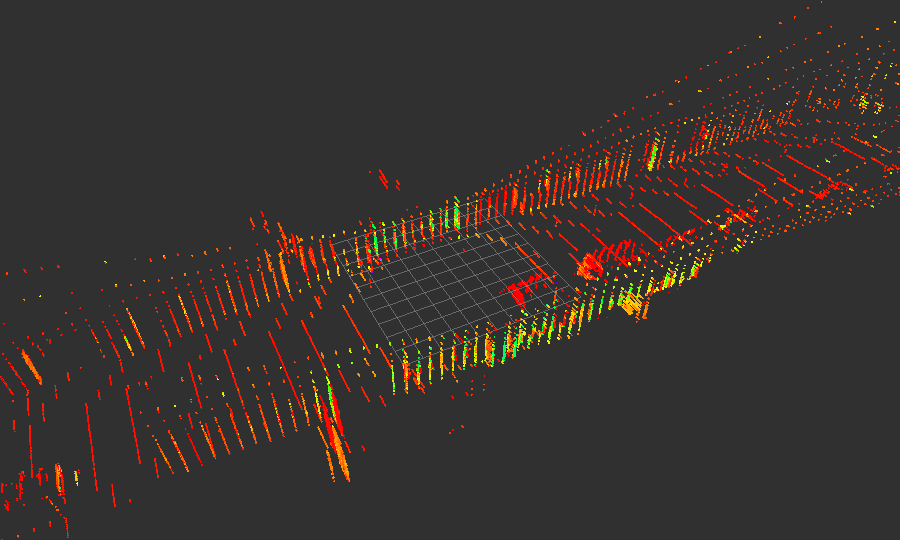}};
    \spy on (4.0,0.3) in node [right] at (-11,4.0);
    \end{tikzpicture}%
    }
    \caption{Google Draco Compression, \\ SNNRMSE=0.1415 m.}\label{fig:distortion2}
  \end{subfigure}
    \begin{subfigure}[t]{0.32\linewidth}
    \resizebox{\linewidth}{!}{%
    \begin{tikzpicture}[spy using outlines={circle,yellow,magnification=2.5,size=5cm, connect spies}]
    \node {\pgfimage{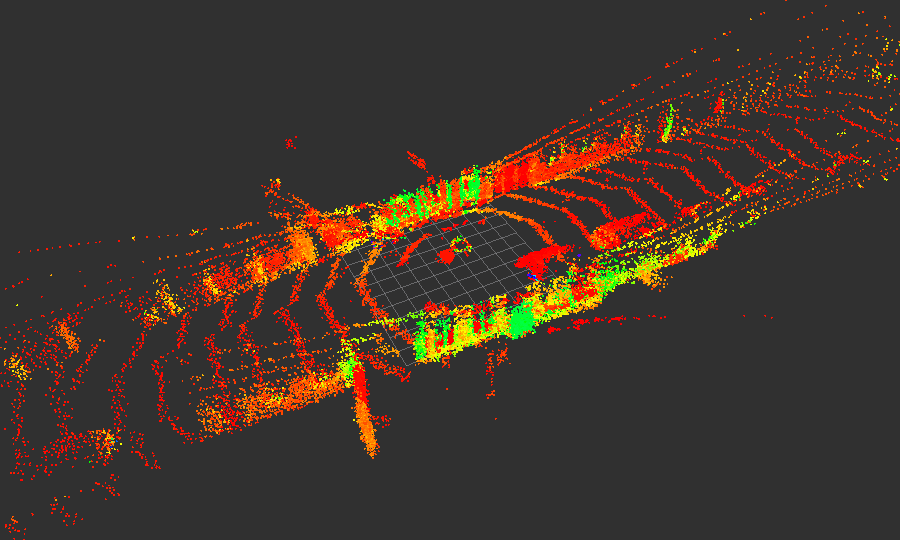}};
    \spy on (4.3,0.85) in node [right] at (-11,4.0);
    \end{tikzpicture}%
    }
    \caption{Tu et al. with JPEG 2000, \\ SNNRMSE=0.1026 m.}\label{fig:distortion3}
  \end{subfigure}
  \begin{subfigure}[t]{0.32\linewidth}
    \resizebox{\linewidth}{!}{%
    \begin{tikzpicture}[spy using outlines={circle,yellow,magnification=2.5,size=5cm, connect spies}]
    \node {\pgfimage{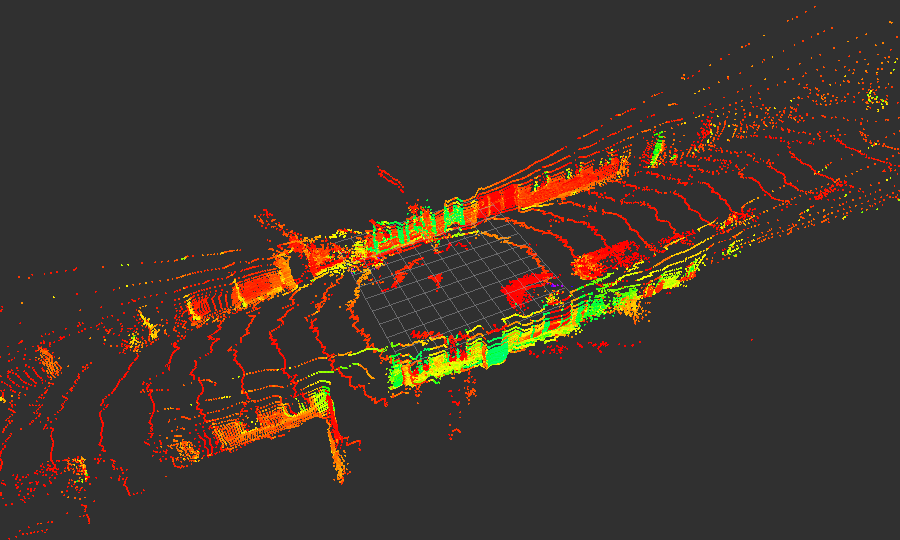}};
    \spy on (4.0,0.25) in node [right] at (-11,4.0);
    \end{tikzpicture}%
    }
    \caption{Proposed method with JPEG2000, \\ SNNRMSE=0.1293 m.}\label{fig:distortion5}
  \end{subfigure}  
  \begin{subfigure}[t]{0.32\linewidth}
    \resizebox{\linewidth}{!}{%
    \begin{tikzpicture}[spy using outlines={circle,yellow,magnification=2.5,size=5cm, connect spies}]
    \node {\pgfimage{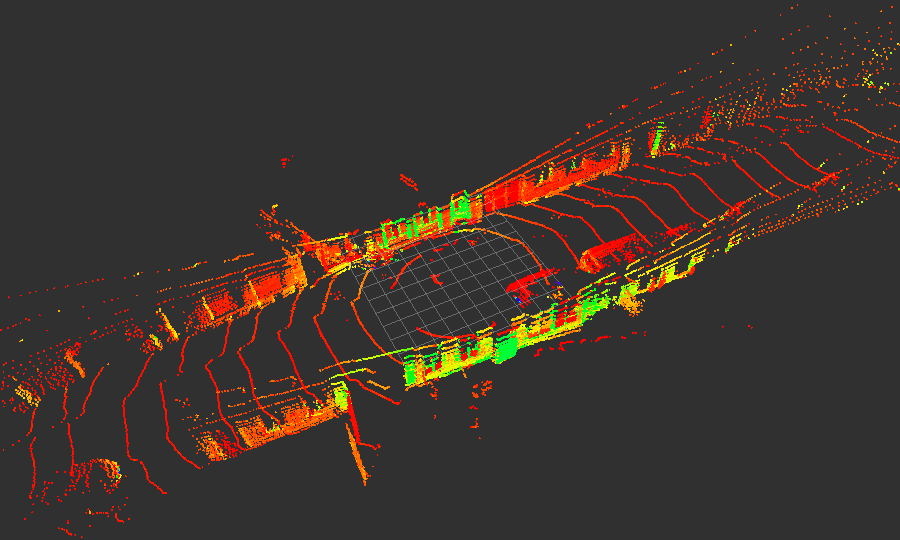}};
    \spy on (4.1,0.3) in node [right] at (-11,4.0);
    \end{tikzpicture}%
    }
    \caption{Proposed method with RNN model, \\ SNNRMSE=0.1207 m.}\label{fig:distortion4}
  \end{subfigure}
  \caption{Visualization of the reconstructed point clouds for different compression methods at similar SNNRMSE error.}
\label{fig:distortion}%
\end{figure*}

Fig. \ref{fig:distortion} shows the compression results at a similar SNNRMSE for the proposed method and reference methods. We can see that the two tree-based methods will overlay points into their cell centers, resulting in a much sparser point cloud. The k-d tree compression has a stronger distortion than Octree compression because the grids in a voxel compress the points into 3D cubics, which quantize the point cloud with the same resolution in three dimensions. However, k-d tree compression segments the points using a 2D plane at each iteration, so the point positions are also quantized onto planes.

The point cloud compression by Tu et al. using JPEG 2000 compression looks noisy, and our proposed method using JPEG 2000 results in distance fluctuations in the radial direction. Our proposed RNN method achieves the best visual reconstruction result without much noticeable distortion or quantization artifacts.
\subsection{Effectiveness}
The evaluations in Fig. \ref{fig:quality} and Fig. \ref{fig:intensitydistortion} were done without compressing the intensity to maintain comparability with other approaches. If we additionally compress the intensity image with JPEG 2000, then the memory footprint can be further reduced by 4 bpp. Hence, we could achieve a total footprint of 10 bpp for the RNN approach which is only 5\% of the original point cloud footprint. If we evaluate the compression performance on basis of the footprint of our three uncompressed image representations, then the compression rate is about 20\%. This comes at the cost of 250ms for the compression step and 150ms for decompression. Using the JPEG 2000 compression instead of RNN compression reduces the compression delay to 35ms and 5ms for decompression.

%% file: sections/5_conclusion.tex
\section{Conclusion}
\label{sec:conclusion}
In this paper, we propose a point cloud compression framework which transforms a 3D point cloud to several 2D matrix representations and then compresses them with image compression methods and a RNN compression approach. In particular, we projected a point cloud losslessly to a combination of range, azimuth, and intensity images. The representations are aligned, calibrated and denoised before the compression step. The evaluation results show that the proposed encoding scheme allows the compression algorithms to better exploit spatial correlations in the data and thus achieve consistent or better point cloud geometric compression quality than comparable approaches. The data transformation also allows for a better compression of the intensity representation. Finally, our RNN model could compress a point cloud to a fraction of its original size while achieving an acceptable reconstruction error, and it also achieved the best visual result compared to all other approaches.

%% file: sections/6_acknowledgment.tex
%